\documentclass[runningheads]{llncs}

\usepackage{multirow}
\usepackage{amsmath}

\usepackage[utf8]{inputenc} 
\usepackage[T1]{fontenc}    
\usepackage{hyperref}       
\usepackage{url}            
\usepackage{booktabs}       
\usepackage{amsfonts}       
\usepackage{nicefrac}       
\usepackage{microtype}      
\usepackage{dsfont}
\usepackage{graphicx}
\usepackage[dvipsnames]{xcolor}
\usepackage{varwidth}
\usepackage[symbol]{footmisc}

\usepackage{hyperref}
\makeatletter
\newcommand{\printfnsymbol}[1]{
  \textsuperscript{\@fnsymbol{#1}}
}
\makeatother
\usepackage{array}
\usepackage[list=true]{subcaption}
\usepackage[noblocks]{authblk}

\begin{document}

\title{Contrastive Learning of Single-Cell Phenotypic Representations for Treatment Classification}

\author{Alexis Perakis\thanks{These authors contributed equally to the paper.}\inst{1}
\and Ali Gorji\printfnsymbol{1}\inst{2}
\and Samriddhi Jain\printfnsymbol{1}\inst{2}
\and Krishna Chaitanya\inst{3}
\and Simone Rizza\inst{2}
\and Ender Konukoglu\inst{3}}

\institute{KOF Swiss Economic Institute, ETH Zurich, Switzerland
\and ETH Zurich, Switzerland
\and Computer Vision Lab, ETH Zurich, Switzerland
}
\maketitle
\begin{abstract}

Learning robust representations to discriminate cell phenotypes based on microscopy images is important for drug discovery.
Drug development efforts typically analyse thousands of cell images to screen for potential treatments.
Early works focus on creating hand-engineered features from these images or learn such features with deep neural networks in a fully or weakly-supervised framework. Both require prior knowledge or labelled datasets. 
Therefore, subsequent works propose unsupervised approaches based on generative models to learn these representations.
Recently, representations learned with self-supervised contrastive loss-based methods have yielded state-of-the-art results on various imaging tasks compared to earlier unsupervised approaches.
In this work, we leverage a contrastive learning framework to learn appropriate representations from single-cell fluorescent microscopy images for the task of Mechanism-of-Action classification.
The proposed work is evaluated on the annotated BBBC021 dataset, and we obtain state-of-the-art results in NSC, NCSB and drop metrics for an unsupervised approach. We observe an improvement of 10\% in NCSB accuracy and 11\% in NSC-NSCB drop over the previously best unsupervised method. 
Moreover, the performance of our unsupervised approach ties with the best supervised approach.
Additionally, we observe that our framework performs well even without post-processing, unlike earlier methods. With this, we conclude that one can learn robust cell representations with contrastive learning.


\end{abstract}

\section{Introduction}
An effective approach in the field of drug discovery is to relate treatments under development to existing ones. Comparing an unknown to a known treatment enables finding desired similarities and avoiding unwanted effects. Cells of interest are exposed to chemical compounds and then imaged using various microscopy techniques. A cell's response mechanism upon treatment is called Mechanism-of-Action (MOA). Such response mechanisms modify a cell's phenotype (morphology) to various degrees. To compare different treatments we can classify them into different MOAs. To perform this classification, we require cell representations that accurately capture the cell morphology. Morphological cell profiling uses image recognition techniques to construct these cell representations. Typically, cell profiling-based MOA classification is done on thousands of images in a transductive learning setting in the literature, where the whole data are included in the training and evaluation process. 

In the literature, many works have focused on automating the tedious task of learning meaningful single-cell representations and utilize them downstream for MOA assignment.
Initial works like~\cite{ljosa2013comparison,singh2014illumination} rely on creating expert engineered features.
Later, deep neural networks are used to learn these features.
Early works making use of neural networks rely on fully supervised approaches~\cite{Kraus2016,godinez2017multi} to learn these features, where~\cite{godinez2017multi} implements multi-scale convolutional neural networks to extract cell morphology directly from images.
Alternatively, ~\cite{caicedo2018weakly} proposes a weakly supervised learning method.
Other works as in~\cite{godinez2018unsupervised,spiegel2019metadata} use the metadata information to devise pseudo-labels for the network supervision.
All the fully or weakly supervised methods above suffer from the requirement of labeled sets, and annotating such images by experts leads to high costs and time-consuming efforts. Even in~\cite{godinez2018unsupervised,spiegel2019metadata}, the labels acquired from metadata can be prone to imprecise labeling due to the nature of the data and treatment analysis techniques.
Some other works use transfer learning techniques~\cite{ando2017improving,pawlowski2016automating}. With these approaches, it may not be feasible to acquire appropriate labeled datasets for pre-training.

These reasons encourage the usage of unsupervised learning approaches 
that provide the following advantages: large unlabeled datasets can be directly used in training, and no pre-training labelled datasets are required. In approaches as above~\cite{caron2018deep,goldsborough2017cytogan,janssens2020fully,lafarge2019}, the aim is to learn robust single-cell representations.
In ~\cite{janssens2020fully} the authors perform clustering on the learned cell representations and use the cluster assignments as labels, where the implementation is similar to~\cite{caron2018deep}.
Alternatively, some works use generative models like variational autoencoders~\cite{lafarge2019} or generative adversarial networks~\cite{goldsborough2017cytogan} to learn such representations.

For representation learning, many recent works propose self-supervised learning methods using unlabeled data. Recently, self-supervised approaches based on a contrastive loss~\cite{hadsell2006dimensionality} have yielded state of the art performance for imaging tasks such as classification, object detection, segmentation~\cite{chen2020simple,he2020momentum,hjelm2018learning,wu2018unsupervised} on benchmark natural image datasets as well as for medical imaging tasks on MR~\cite{chaitanya2020contrastive}, CT~\cite{xie2020pgl,yan2020self}, X-ray~\cite{azizi2021big,sriram2021covid,vu2021contrastive,zhang2020contrastive}, dermatology~\cite{azizi2021big}, electron microscopy~\cite{Huang2020} image datasets and ECG signals~\cite{kiyasseh2020clocs}.
These contrastive loss based approaches outperform traditional self-supervision based pretext tasks (e.g., rotation~\cite{gidaris2018unsupervised}, inpainting~\cite{pathak2016context}) and generative models~\cite{donahue2016adversarial,donahue2019large}.

In cell profiling, the datasets contain thousands or millions of images.
Unsupervised methods are promising and provide more viable solutions for such applications than supervised methods. Hence, in the proposed work, we leverage the popular contrastive learning framework~\cite{chen2020simple} for learning robust cell representations using only unlabeled data. In contrastive loss-based learning~\cite{hadsell2006dimensionality}, representations are learned by contrasting positive examples to negative examples. As in~\cite{chen2020simple}, we train the network to pull the representations of positive examples to be close in the latent space and push the negative examples representations to be far away from positive examples. 

Our contributions are:
\begin{itemize}
    \item We are the first to learn single-cell representations using a contrastive learning framework in an unsupervised setting.
    \item For a cell profiling dataset, we evaluate and find the most important components and hyper-parameters used in the contrastive framework such as: (i) encoder network size, (ii) data augmentation strategies, (iii) projection head size, (iv) batch size value and (v) temperature parameter.
    \item We achieve state-of-the-art NSCB accuracy and NSC-NSCB drop for an unsupervised method. Our unsupervised results in NSC, NSCB, and drop metrics match the state-of-the-art for a supervised method. 
    \item The learned single-cell representations perform well even without any post-processing, unlike earlier works.
\end{itemize}


\section{Methods}

We divide this section into two parts: (a) representation learning and (b) MOA classification.
In the first part, we learn representations in a contrastive learning framework as proposed in~\cite{chen2020simple}. 
In the second part, we use these representations for the downstream task of Mechanism-of-Action classification of the treatment profiles.

\begin{figure}[!t]
\centering
\includegraphics[width=\textwidth]{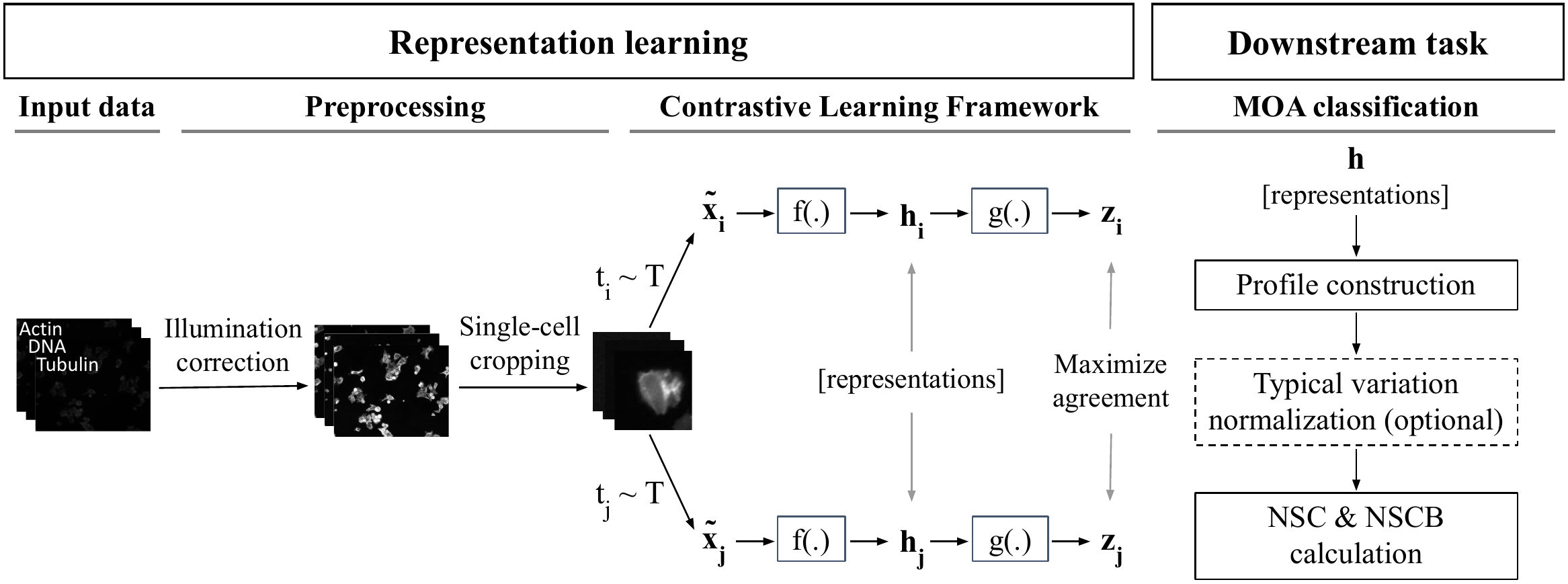}
\caption{Our experimental pipeline can be split into two parts: representation learning and downstream task. In representation learning, we feed single-cell images into the contrastive learning framework~\cite{chen2020simple}, where single-cell representations are extracted. Subsequently, these representations are aggregated into treatment profiles and post-processed (optional step) in the downstream task. We use the resulting profiles for MOA classification and calculate NSC and NSCB accuracies.}
\label{fig:main_method_figure}
\end{figure}

\textbf{(a) Representation learning:} In the contrastive framework, we learn a global representation for each input image as illustrated in Fig.~\ref{fig:main_method_figure}. We follow~\cite{chen2020simple} and sample a mini-batch of images of size $N$ from the whole dataset $X$. Then, for each sampled image $x$ we apply two random transformations $t_i$ and $t_j$ (sampled from a set of transformations $T$). The transformed images are denoted by $\tilde{x}_i = t_i(x)$ and $\tilde{x}_j = t_j(x)$ to obtain $2N$ images in the batch, as shown in Fig.~\ref{fig:main_method_figure}.
For the encoder network, we use a ResNet~\cite{he2016deep} denoted by $f$ to get the representation $h$ where $h_i = f(x_i)$, which is followed by a projection head $g$. The output latent representation is given by $z$ where $z_i=g(f(x_i))$. 
The two transformed images arising from a given image $x$ are denoted as the positive pair. The remaining $2(N-1)$ images in the batch act as the negative pairs and form the negative images set $\Omega^-$.
The contrastive loss applied on the positive pairs of output latent representations is defined as follows:
\begin{equation}
    l(\tilde{x}_i, \tilde{x}_j)= - \log \frac{e^{\mathrm{sim}(z_i, z_j)/\tau}}{e^{\mathrm{sim}(z_i, z_j)/\tau} + \sum_{x_n \in \Omega^-} e^{\mathrm{sim}(z_i, {{g}}({{f}}(x_n)))/\tau}}
    \label{eq:global_pair_loss}
\end{equation}
where $z_i,z_j$ are the output latent representations of the positive pair ($x_i,x_j$) and $x_n$ are the corresponding negative images from the set $\Omega^-$. $\tau$ denotes the temperature parameter. The similarity between two representations is computed using cosine similarity, which is defined as $\mathrm{sim}(z_{i},z_{j}) = z_{i}^{T}z_{j}/\|z_{i}\|\|z_{j}\|$. 

\noindent The net contrastive loss across all positive pairs in the batch is given below:
\begin{equation}
    L_{net} = \frac{1}{2N} \sum_{k=1}^{N} [l(t_i(x_k), t_j(x_k)) + l(t_j(x_k), t_i(x_k))]
    \label{eq:global_net_loss}
\end{equation}
By optimizing the loss $L_{net}$, we enable the network to learn representations of a positive pair for a given image to be similar under different transformations such as crop, rotation, color jitter, etc. Also, they should be dissimilar to representations of the remaining images in the batch that constitute the negative set.
With this optimization, we aim to learn robust and meaningful global representations $h=f(.)$ that can be used for downstream tasks.

\textbf{(b) MOA Classification:} Our proposed method is evaluated on a downstream MOA classification task.  
The final latent representations $z_i$ and projection head $g(.)$ are only used for the representation learning stage and are discarded for the classification step.
The Mechanism-of-Action classification is performed on the learned cell representations $h_i$ that are output from the encoder network.

We use two widely applied procedures for evaluating how MOA are assigned to treatments: not-same-compound (NSC) matching from \cite{ljosa2013comparison} and not-same-compound-and-batch (NSCB) matching from \cite{ando2017improving}.
In a first step, single-cell representations are aggregated into \textit{treatment profiles} as described in \cite{ljosa2013comparison}.
This aggregation results in a treatment profile vector for each treatment in the dataset.

In the second step, one MOA is assigned to each treatment profile. 
Both NSC and NSCB scores are $1^{st}$ nearest neighbour MOA classification accuracies where the distance measure used is cosine distance.
Treatments are assigned an MOA label one at a time. In NSC matching, for a given treatment, we search the representation space for the nearest neighbour not containing the same compound. Recall that a treatment is a compound-concentration pair. In NSCB matching, the search further excludes all treatments from the same experimental batch.

\section{Datasets and Network details}
\textbf{Dataset:} Our work uses the annotated part of the BBBC021 dataset~\cite{caie2010high} available from the Broad Bioimage Benchmark Collection~\cite{bbbc2012}. BBBC021 consists of multi-channel images captured from human MCF-7 breast cancer cells exposed to chemical compounds for 24 hours. Cells are imaged by fluorescent microscopy. 
The 3 grey-scale channels represent DNA, B-tubulin and F-actin. The proposed method is evaluated on the subset of BBBC021 that has previously been labelled for MOAs by \cite{ljosa2013comparison}. There are 12 distinct MOA present in this subset for a total of 103 treatments and 38 compounds. Each treatment corresponds to a compound-concentration combination. The control cells are treated with DMSO (dimethyl sulfoxide). We solely use the ground truth annotated part of BBBC021 for our study, in line with the evaluation strategy applied in all the works in the literature~\cite{ando2017improving,caicedo2018weakly,godinez2017multi,godinez2018unsupervised,goldsborough2017cytogan,janssens2020fully,Kraus2016,lafarge2019,ljosa2013comparison,pawlowski2016automating,singh2014illumination,Tabak2019}.

\noindent\textbf{Data Pre-processing:}
We apply the illumination correction algorithm \cite{singh2014illumination} on the original images as performed in \cite{caicedo2018weakly}.
Then, we crop all the single-cell instances from these images to fixed image dimensions of $96 \times 96$, based on the cell locations introduced by \cite{ljosa2013comparison}.

\noindent\textbf{Post-processing:} We use
typical variation normalization (TVN) \cite{caicedo2018weakly,janssens2020fully,lafarge2019,Tabak2019} to reduce batch effects and improve profiles. It comprises of a \textit{whitening} step and a \textit{correlation alignment} (CORAL) \cite{Sun2016} step. We also evaluate the effects of using only whitening~\cite{janssens2020fully,lafarge2019,Tabak2019} as well as no post-processing. 

\noindent\textbf{Network details:}
We choose ResNet~\cite{he2016deep} for the encoder network $f(.)$ and MLP layers for the projection head network $g(.)$.

\noindent\textbf{Training details:}
We train the network with the following parameter settings. We choose Adam as the optimizer with a learning rate of $3e^{-4}$ and a weight decay of $1e^{-5}$. 
We run the initial hyper-parameter evaluation for 150 epochs on a subset of 15\% of the annotated images. We run the final experiments with the optimal set of hyper-parameters for up to 600 epochs on the complete set of annotated data. All our experiments are performed on a Nvidia Titan Xp GPU.

\section{Experiments}\label{experiments}
\noindent\textbf{Experimental Setup:}
The part of BBBC021 annotated for MOAs contains 2'526 original images which translates into 454'793 single-cells. First, we tune our hyperparameters on a subset of the annotated BBBC021 dataset containing 15\% of the images as performed by~\cite{Kraus2016}. This significantly reduces our training time and compute resources. In a second step, we train our framework on the whole annotated BBBC021 dataset and report our final results. Note that we perform training and evaluation in a transductive learning setting on the whole annotated data in line with the evaluation strategy used in the earlier works. 

\noindent\textbf{Evaluation:} We use NSC, NSCB scores and the NSC-NSCB drop to measure the performance.

\noindent\textbf{Ablation study of hyper-parameters:}
For the ablation study of hyper-parameters, we only use 15\% of the annotated images. We report all results for these experiments with TVN post-processing. (Refer to the Supplementary for results using only whitening or without post-processing.) We evaluate the following hyper-parameters to analyze their effect on downstream performance.

(i) \textbf{Encoder Network Size $f(.)$}: For the encoder, the following ResNet sizes are evaluated: ResNet18, ResNet50, ResNet101. The remaining hyper-parameters are investigated for a ResNet50 encoder as it yielded the best results.

(ii) \textbf{Data Augmentation strategy}: We explore different data augmentations ($T$) such as crop, flip, rotations by 90 degrees, color jitter, grey-distortion, and Gaussian blur. In the default setting, we have all augmentations. We experiment by removing one augmentation at a time from this main set of augmentations to analyze the importance of each.

(iii) \textbf{Projection head $g(.)$}: We explore three types of projection heads: the identity, a linear projection and a non-linear projection (two layer MLP).

(iv) \textbf{Batch Size}: We evaluate the following batch sizes used for each training iteration: 64, 128, 256.

(v) \textbf{Temperature coefficient}: As done in~\cite{chen2020simple}, we evaluate the effect of temperature for the following values: $\tau= {0.05, 0.1, 0.5, 1}$ for different combinations of other hyper-parameters.

\noindent\textbf{Final experiments and comparison:}
We choose the hyper-parameters for the final set of experiments from the ablation results. They are a ResNet50 as the encoder network, a two-layer MLP as the projection head, an augmentation strategy without the grey distortion, a batch size of 256 and a temperature parameter of $\tau = 0.5$. Here, the evaluation is performed in a transductive learning setting on the whole annotated dataset as described earlier.

\noindent \textbf{TVN post-processing}: We also evaluate if there is any difference in performance with and without TVN post-processing applied to the learned representations.

\section{Results}~\label{results}
\textbf{Ablation study of hyper-parameters:}

(i) Encoder size: In Table~\ref{Enc_Aug_withTVN}, we present the results for different ResNet architectures evaluated as the encoder network. We observe that ResNet18 slightly outperforms both ResNet50 and ResNet101.
However, when we compare ResNet50 to ResNet18, we observe smaller values for NSC-NSCB drop with and without post-processing for ResNet50 (Refer to Table 4 in Supplementary). Hence, we choose ResNet50 for the remaining hyper-parameter evaluation.

\begin{table}[!t]\centering
\caption{Model performance is shown for different encoder network sizes ((a) $f$ size) and for one augmentation removed at a time ((b) Augmentation removal) under TVN post-processing conditions. Jitter, Grey, and Blur refer to color-jitter, Grey color distortion, Gaussian Blur, respectively.} 

\begin{tabular}{c c c c c c c c c c c} \toprule
    \multicolumn{1}{l}{Metric} &
    \multicolumn{3}{c}{(a) $f$ size} &
    \multicolumn{7}{c}{(b) Augmentation removal} \\
    \cmidrule(lr){2-4}
    \cmidrule(lr){5-11}
    
    &
    \multicolumn{1}{c}{ResNet18} & \multicolumn{1}{c}{ResNet50} & \multicolumn{1}{c}{ResNet101} &
    \multicolumn{1}{c}{None} & \multicolumn{1}{c}{Crop} & \multicolumn{1}{c}{Flip} & \multicolumn{1}{c}{Rotation} & \multicolumn{1}{c}{Jitter} & \multicolumn{1}{c}{Grey} & \multicolumn{1}{c}{Blur}\\ 
    \midrule
    NSC & $95\%$ & $93\%$ &$92\%$ &
     $94\%$ & $83\%$ & $93\%$ & $94\%$  & $91\%$ &$96\%$ &$92\%$\\
    NSCB & $91\%$ & $91\%$ & $90\%$ &
    $91\%$ & $70\%$ & $88\%$ & $87\%$ & $88\%$  & $94\%$ &$88\%$\\
    \bottomrule
\end{tabular}
\label{Enc_Aug_withTVN}
\end{table}

(ii) Data augmentation strategies: In Table~\ref{Enc_Aug_withTVN}, we present our results for the different augmentation strategies adopted. We drop one augmentation at a time to analyze its importance. We observe that dropping the grey color distortion significantly improves NSC and NSCB performance. This could be due to grey-scale averaging over the three channels. It may not be meaningful for such datasets as they contain relatively independent pixel intensities in each channel, with different channels capturing different parts of a cell. Typical RGB channels contain dependent information.
The most important augmentation was found to be cropping followed by color jitter. This is also observed in earlier works~\cite{chen2020simple} on natural images.

Other hyper-parameter evaluations are presented in the Supplementary Material in Tables 5, 6 and 7. We summarize our findings to be the following:
(iii) a two-layer MLP (non-linear projection) and linear projection yield similar NSC and NSCB scores. They both yield higher scores than the identity. 
(iv) A batch size of 256 yields higher NSC and NSCB scores on average over batch sizes of 64 and 128.
(v) The temperature coefficients of 0.1 and 0.5 yield higher scores compared to a large value of 1 and very small value of 0.05.

\noindent \textbf{Final experiments and comparison:}
In Table~\ref{tab:main}, we present our results from the final set of experiments on the whole annotated dataset in a transductive setting as evaluated in earlier works. We observe that the proposed unsupervised contrastive learning framework yields better results than earlier unsupervised works with an improvement over the best method~\cite{janssens2020fully} of $10\%$ and $11\%$ in NSCB scores and drop respectively.
We also observe that the proposed framework yields similar scores to supervised counterparts~\cite{ando2017improving} where large number of annotations or suitable pre-training datasets are required to achieve such high performance.

\begin{table}[!b]
\vspace{-0.5cm}
\centering
\caption{MOA classification accuracy. * indicates results without post-processing.}
\begin{tabular}{lccc|lcccc}\toprule
    \multicolumn{4}{c|}{Supervised} &
    \multicolumn{4}{c}{Unsupervised}\\
    \cmidrule(lr){1-4}
    \cmidrule(lr){5-8}

    Method & NSC & NSCB & Drop & Method & NSC & NSCB & Drop \\ \midrule
    Ljosa et al. \cite{ljosa2013comparison} & $94\%$ & $77\%$ & $17\%$ &  Janssens et al. \cite{janssens2020fully} & $\mathbf{97\%}$ & $\mathbf{85\%}$ & $12\%$ \\
    Singh et al. \cite{singh2014illumination} & $90\%$ & $85\%$ & NA & Lafarge et al. \cite{lafarge2019} & $93\%$ & $82\%$ & $\mathbf{11\%}$ \\ 
    Ando et al. \cite{ando2017improving} & $\mathbf{96\%}$ & $\mathbf{95\%}$ & $\mathbf{1\%}$ & Lafarge et al. \cite{lafarge2019} * & $92\%$ & $72\%$ & $20\%$ \\
    Pawlowski et al. \cite{pawlowski2016automating} & $91\%$ & NA & NA &  Our work * & $\mathbf{97}\%$ & $94\%$ & $3\%$ \\
    Caicedo et al.~\cite{caicedo2018weakly} & $95\%$ & $89\%$ & $6\%$& Our work + whitening & $\mathbf{96\%}$ & $\mathbf{95\%}$ & $\mathbf{1\%}$ \\ 
     &  &  &  & Our work + TVN & $\mathbf{97\%}$ & $92\%$ & $5\%$ \\
    
\bottomrule
\end{tabular}

\label{tab:main}
\end{table}

\noindent \textbf{TVN post-processing:}
We observe that our method performs well even without applying post-processing such as TVN or whitening. 
Earlier works are sensitive to the post-processing step where a 10\% decrease in NSCB is observed for~\cite{lafarge2019}. Our work has only a 1\% decrease in NSCB when removing the whitening step, as shown in Table~\ref{tab:main}.

\begin{figure}
\centering

\includegraphics[width=0.78\linewidth]{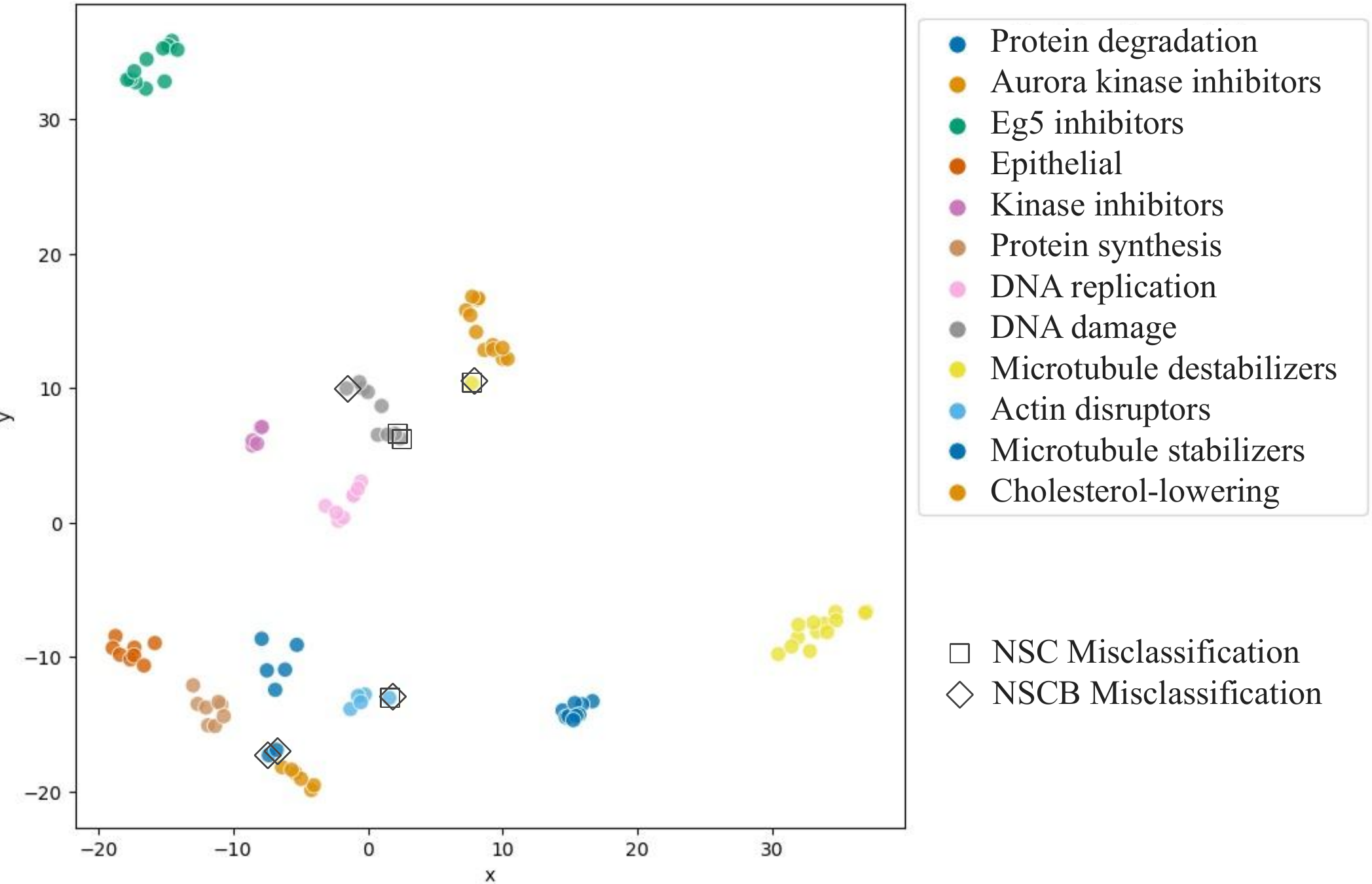}
\label{MOAstatus}
\caption{t-SNE plot for treatment profiles subject to whitening post-processing.}
\label{tSnePlots}

\end{figure}

\noindent\textbf{t-SNE plot:} Figure~\ref{tSnePlots} shows a t-SNE visualization of the treatment profiles obtained with our final experiment. Here, the treatments are classified into the 12 MOAs available. In Figure~\ref{tSnePlots}, we can observe that only 4 out of 103 and 5 out of 92 treatments are classified incorrectly during NSC and NSCB MOA assignment respectively. NSC and NSCB mis-classifications are marked with black squares and black diamonds respectively. 

\section{Conclusion}~\label{conclusion}
In this work, we conclude that using an unsupervised approach with contrastive learning on single-cell images leads to excellent representations in morphological cell profiling. With this framework, we demonstrate state of the art results in an unsupervised learning setting for the downstream task of MOA classification. 
In the unsupervised setting, our NSCB score of $95\%$ for MOA matching is the highest classification accuracy reported and similarly the drop of $1\%$ between our NSC and NSCB scores is the best reported. Furthermore, our unsupervised results are identical to the state of the art transfer learning approach~\cite{ando2017improving} and higher than the supervised approach~\cite{caicedo2018weakly}, both relying on labels. 
We perform an ablation study of hyper-parameters and conclude that encoder size as well as data augmentations are the most crucial hyper-parameters for obtaining maximum improvements on cell profiling datasets.
Finally, we also show that the performance of the resulting representations does not deteriorate even when removing post-processing techniques.

\newpage
\bibliographystyle{splncs04}
\bibliography{references}

\newpage

\section*{Supplementary}

\subsection*{Ablation study results}

\begin{table}[h]\centering
\caption{Model performance for \textbf{different augmentation sets by removing one augmentation at a time.} 
"None" refers to using all the augmentations. "Blur" refers to Gaussian Blur.
We perform all experiments with a ResNet50 encoder network, a two-layer MLP ($2048\times256$, ReLU) as projection head, a batch size of 256, and train for 150 epochs. }
\begin{tabular}{l l c c c c c c c} \toprule
    \multicolumn{1}{l}{Post-processing} &
    \multicolumn{1}{l}{Metric} &
    \multicolumn{7}{c}{Augmentation removal} \\
    \cmidrule(lr){3-9}
    
    & & 
    \multicolumn{1}{c}{None} & \multicolumn{1}{c}{Crop} & \multicolumn{1}{c}{Flip} & \multicolumn{1}{c}{Rotation} & \multicolumn{1}{c}{Color-Jitter} & \multicolumn{1}{c}{Grey-Distortion} & \multicolumn{1}{c}{Blur}\\ 
    \midrule
    \multirow{2}{*}{None} & NSC & $88.3\%$ & $81.6\%$ &$89.3\%$ &$86.4\%$ &$87.4\%$ &$94.2\%$ &$88.3\%$  \\
    & NSCB & $85.9\%$ & $67.4\%$ & $82.6\%$ &$79.3\%$ &$77.2\%$ &$91.3\%$ &$84.8\%$ \\
    \midrule
    \multirow{2}{*}{Whitening} & NSC & $94.2\%$ & $82.5\%$ &$95.1\%$ &$94.2\%$ &$93.2\%$ &$96.1\%$ &$94.2\%$ \\
    & NSCB & $91.3\%$ & $70.7\%$ & $89.1\%$ &$89.1\%$ &$89.1\%$ &$90.2\%$ &$89.1\%$ \\
    \midrule
    \multirow{2}{*}{TVN} & NSC & $94.2\%$ & $82.5\%$ &$93.2\%$ &$94.2\%$ &$91.3\%$ &$96.1\%$ &$92.2\%$\\
    & NSCB & $91.3\%$ & $69.6\%$ & $88.0\%$ &$87.0\%$ &$88.0\%$ &$93.5\%$ &$88.0\%$ \\
\bottomrule
\end{tabular}
\label{appendix:augmentation}
\vspace{-0.75cm}
\end{table}

\begin{table}[h]\centering
\caption{Model performance for \textbf{different ResNet sizes} as the $f(.)$ network. We perform all experiments with the augmentation set containing all augmentations except grey-distortion, a two-layer MLP ($2048\times256$, ReLU) as projection head, a batch size of 128, and train for 150 epochs.}
\begin{tabular}{l l c c c} \toprule
    Post-processing & Metric & ResNet18 & ResNet50 & ResNet101\\ \midrule
    \multirow{3}{*}{None} & NSC & $93.2\%$ & $93.2\%$ &$86.4\%$  \\
    & NSCB & $87.0\%$ & $89.1\%$ & $89.1\%$  \\
    & drop(NSC-NSCB) & $6.2\%$ & $4.1\%$ & $-2.7\%$  \\
    \midrule
    \multirow{3}{*}{Whitening} & NSC & $94.2\%$ & $93.2\%$ &$92.2\%$  \\
    & NSCB & $91.3\%$ & $90.2\%$ & $90.2\%$  \\
    & drop(NSC-NSCB) & $2.9\%$ & $3.0\%$ & $2.0\%$  \\
    \midrule
    \multirow{3}{*}{TVN} & NSC & $95.1\%$ & $93.2\%$ &$92.2\%$  \\
    & NSCB & $91.3\%$ & $91.3\%$ & $90.2\%$  \\
    & drop(NSC-NSCB) & $3.8\%$ & $1.9\%$ & $2.0\%$  \\
    
\bottomrule
\end{tabular}
\label{appendix:resnet}
\vspace{-0.75cm}
\end{table}

\begin{table}\centering
\caption{Model performance for \textbf{different batch sizes}. We perform all experiments with a ResNet50 encoder network, with the augmentation set containing all augmentations except grey-distortion, a two-layer MLP (2048x256, ReLU) as projection head, and train for 150 epochs.}
\begin{tabular}{l l c c c} \toprule
    Post-processing & Metric & $Batch Size = 64$ & $Batch Size = 128$ & $Batch Size = 256$\\ \midrule
    \multirow{2}{*}{None} & NSC & $94.2\%$ & $89.3\%$ &$94.2\%$  \\
    & NSCB & $89.1\%$ & $89.1\%$ & $91.3\%$  \\
    \midrule
    \multirow{2}{*}{Whitening} & NSC & $94.2\%$ & $93.2\%$ &$96.1\%$  \\
    & NSCB & $88.0\%$ & $90.2\%$ & $90.2\%$  \\
    \midrule
    \multirow{2}{*}{TVN} & NSC & $93.2\%$ & $93.2\%$ &$96.1\%$  \\
    & NSCB & $88.0\%$ & $91.3\%$ & $93.5\%$  \\
\bottomrule
\end{tabular}
\label{appendix:batch}
\vspace{-0.8cm}
\end{table}

\begin{table}\centering
\caption{Model performance for \textbf{different projection head $g(.)$ networks}. We perform all experiments with a ResNet18 encoder network, with the augmentation set containing all augmentations except grey-distortion, a batch size of 128, and train for 150 epochs. Due to lack of higher memory GPUs, we perform this evaluation on ResNet18 encoder instead of ResNet50.}
\begin{tabular}{l l c c c} \toprule
    Post-processing & Metric & Identity & Linear & Non-linear (2-layer MLP)\\ \midrule
    \multirow{2}{*}{None} & NSC & $81.6\%$ & $92.2\%$ &$93.2\%$  \\
    & NSCB & $69.6\%$ & $82.6\%$ & $87.0\%$  \\
    \midrule
    \multirow{2}{*}{Whitening} & NSC & $84.5\%$ & $93.2\%$ &$94.2\%$  \\
    & NSCB & $70.7\%$ & $89.1\%$ & $91.3\%$  \\
    \midrule
    \multirow{2}{*}{TVN} & NSC & $83.5\%$ & $95.1\%$ &$95.1\%$  \\
    & NSCB & $69.6\%$ & $91.3\%$ & $91.3\%$  \\
\bottomrule
\end{tabular}
\label{appendix:projection}
\vspace{-0.8cm}
\end{table}

\begin{table}\centering
\caption{Model performance for \textbf{different values of the temperature parameter $\tau$}. We perform all experiments with a ResNet50 encoder network, with the augmentation set containing all augmentations except grey-distortion, a two-layer MLP (2048x256, ReLU) as projection head, a batch size of 256, and train for 150 epochs.}
\begin{tabular}{l l c c c c} \toprule
    Post-processing & Metric &  $\tau=1$ & $\tau=0.5$ & $\tau=0.1$ & $\tau=0.05$ \\ \midrule
    \multirow{2}{*}{None} & NSC & $81.6\%$ & $93.2\%$ &$94.2\%$ & $91.3\%$  \\
    & NSCB & $83.7\%$ & $88.0\%$ & $91.3\%$ & $81.5\%$ \\
    \midrule
    \multirow{2}{*}{Whitening} & NSC & $87.4\%$ & $93.2\%$ &$96.1\%$ & $91.3\%$ \\
    & NSCB & $89.1\%$ & $88.0\%$ & $90.2\%$ & $87.0\%$ \\
    \midrule
    \multirow{2}{*}{TVN} & NSC & $86.4\%$ & $93.2\%$ &$96.1\%$ & $93.2\%$ \\
    & NSCB & $88.0\%$ & $91.3\%$ & $93.5\%$ & $85.9\%$ \\
\bottomrule
\end{tabular}
\label{appendix:temp}
\end{table}

\end{document}